\documentclass[final]{cvpr}

\usepackage{times}
\usepackage{epsfig}
\usepackage{graphicx}
\usepackage{amsmath}
\usepackage{amssymb}
\usepackage{mathrsfs}
\usepackage{makecell}
\usepackage{booktabs}
\usepackage{multirow}
\usepackage{caption}
\usepackage{algorithmicx}  
\usepackage{algorithm}
\usepackage{algpseudocode}  
\usepackage{enumitem}
\usepackage{subfig}

\usepackage[pagebackref=true,breaklinks=true,colorlinks,bookmarks=false]{hyperref}

\newcommand{\tabincell}[2]{\begin{tabular}{@{}#1@{}}#2\end{tabular}} 

\def\cvprPaperID{1987} 
\def\confYear{CVPR 2021}

\begin{document}

\title{Data-Uncertainty Guided Multi-Phase Learning for \\Semi-Supervised Object Detection}

\author{
Zhenyu Wang \qquad Yali Li \thanks{Corresponding author} \qquad Ye Guo \qquad Lu Fang \qquad Shengjin Wang\\



Department of Electronic Engineering, Tsinghua University\\
{\tt\small \{wangzy20, guo-y18\}@mails.tsinghua.edu.cn,  \{liyali13, fanglu, wgsgj\}@tsinghua.edu.cn}

}

\maketitle

\pagestyle{empty}
\thispagestyle{empty}

\begin{abstract}
    In this paper, we delve into semi-supervised object detection where unlabeled images are leveraged to break through the upper bound of fully-supervised object detection. Previous semi-supervised methods based on pseudo labels are severely degenerated by noise and prone to overfit to noisy labels, thus are deficient in learning different unlabeled knowledge well. To address this issue, we propose a data-uncertainty guided multi-phase learning method for semi-supervised object detection. We comprehensively consider divergent types of unlabeled images according to their difficulty levels, utilize them in different phases, and ensemble models from different phases together to generate ultimate results. Image uncertainty guided easy data selection and region uncertainty guided RoI Re-weighting are involved in multi-phase learning and enable the detector to concentrate on more certain knowledge. Through extensive experiments on PASCAL VOC and MS COCO, we demonstrate that our method behaves extraordinarily compared to baseline approaches and outperforms them by a large margin, more than $3\%$ on VOC and $2\%$ on COCO.
\end{abstract}

\vspace{-1.0em}

\section{Introduction}

With the success of Convolutional Neural Networks (CNNs) \cite{krizhevsky2012imagenet, lecun1998gradient}, object detection methods have been promoted rapidly in recent years. Plenty of object detection models  \cite{girshick2014rich, girshick2015fast, ren2015faster, redmon2016you, liu2016ssd} achieve superior performance on benchmark datasets \cite{everingham2010pascal, lin2014microsoft}. However, these models are heavily dependent on a large amount of fully-supervised data with complete category and bounding box annotations, which are labor-intensive to collect \cite{kuznetsova2018open}. 

\begin{figure}[]
      \subfloat[]{ \includegraphics[width=0.98\columnwidth]{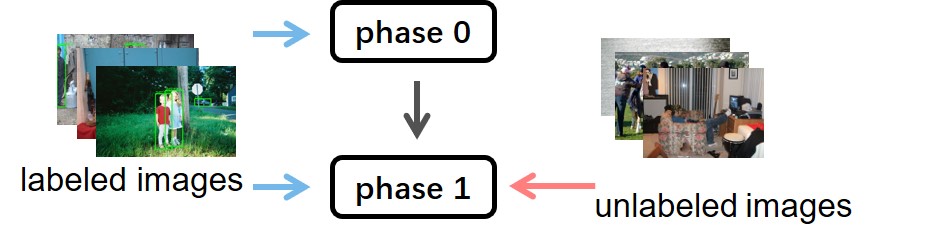} \label{fig:frame1}}\\
      \subfloat[]{ \includegraphics[width=0.98\columnwidth]{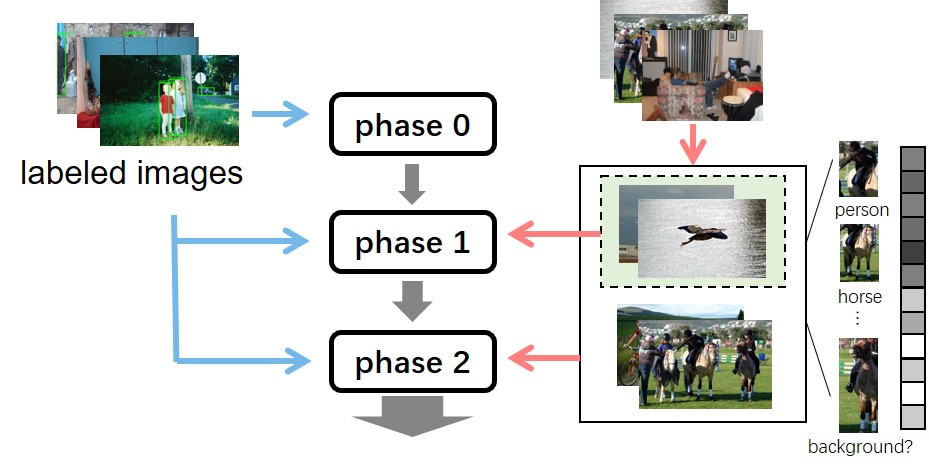} \label{fig:frame2}} \\
\caption{\textbf{Frameworks for previous pseudo-label based one-phase training (a) and our multi-phase method (b).} Image-level uncertainty based selection and region-level uncertainty based re-weighting guide our multi-phase learning for handling noise in pseudo labels.}
\label{fig:frame}
\end{figure}

To address the preceding problems, semi-supervised object detection (SSOD) \cite{rosenberg2005semi} receives much attention recently. It exploits a large amount of unlabeled data to boost the performance of fully-supervised object detection, especially when only limited labeled data are available. Currently, many SSOD methods \cite{radosavovic2018data} are established on pseudo labels \cite{lee2013pseudo} and adopt the one-phase learning scheme in Fig. \ref{fig:frame1}. With a pre-trained fully-supervised model on labeled images, pseudo annotations of unlabeled images are obtained. These pseudo labels are treated as groundtruth of unlabeled data and are integrated with annotations of labeled data to train the SSOD model.

Despite that the one-phase learning is somewhat effective for SSOD, it is insufficient for knowledge excavation to exploit the unlabeled data only once. The reason lies in the fact that noise is inherently attached to pseudo annotations. Deep learning based models have the potential to fit any training annotation, even the incorrect ones. When pseudo annotations are noisy with some false information, detection models are also able to "learn" to fit them. This fitting ability to incorrect annotations surpasses the representative learning for correct ones. We call this phenomenon as \textit{label noise overfitting} problem, which is also corroborated in previous studies \cite{arpit2017closer, zhang2016understanding, ma2018dimensionality}. As a result, SSOD models with one-phase learning tend to fit the \textit{difficult} data with more noise, while neglecting the \textit{easy} data with high confidence.

The negative impacts of label noise overfitting are mainly two-fold. On the one hand, at \emph{image level}, difficult images with much noise preponderate during training, making the detector inflexible to employ unlabeled data with different difficulty levels. On the other hand, at \emph{region level}, regions overlapped heavily with existing objects but lacking pseudo annotations incorporate much more noisy information and dominate the training, contributing extensive incorrect gradient messages to SSOD training.

To tackle this, we describe the noisy labeled data with uncertainty and propose a data-uncertainty guided multi-phase learning for SSOD. At image level, we introduce the uncertainty to guide the image selection for different phases of training. In practice, we perform SSOD training on \textit{easy} unlabeled images with low uncertainty first, then continue with \textit{difficult} unlabeled images with high uncertainty. During this process, we collect more than one model which experts in images with different difficulty levels separately. They cooperatively specialize all unlabeled images so we aggregate them together to complement each other for inference. At region level, we measure the uncertainty of background regions based on their similarity and overlaps with each other. We further conduct RoI re-weighting by the guidance of region uncertainty degrees and involve it in the multi-phase training. This RoI re-weighting strategy reduces weights for uncertain regions and forces detection models to pay more attention to certain regions.

Our main contributions can be summarized as follows:

\begin{itemize}[topsep=3pt, parsep=0pt, itemsep=3pt, partopsep=0pt]
    \item We propose an uncertainty guided multi-phase SSOD learning method. With image-uncertainty based selection, we alleviate attention imbalance on different difficulty levels of data and are capable of fitting all unlabeled images well.
    \item We introduce a region-uncertainty based RoI re-weighting strategy to guide multi-phase learning and assist the detector in focusing on more certain regions.
    \item On the PASCAL VOC and MS COCO dataset, our method reaches 78.6\% and 42.3\%, which exceeds the state-of-the-art by 2.4\%, 2.2\%, respectively.
\end{itemize}

\section{Related Work}

\textbf{Object detection} is one of the most important tasks in computer vision. It aims to detect objects from an image, predict correct classification categories, and assign accurate bounding boxes. It is generally divided into two-stage detection methods and one-stage methods. Two-stage detectors \cite{girshick2014rich, girshick2015fast, ren2015faster, dai2016r, he2017mask} usually produce region proposals then perform classification and regression on these proposals, while one-stage detectors \cite{redmon2016you, liu2016ssd, redmon2018yolov3} generate bounding boxes prediction and region classification directly. Fully-supervised object detection (FSOD) develops rapidly recently \cite{lin2017focal, shrivastava2016training, Cai_2019, DynamicRCNN} and achieves outstanding results in benchmark datasets \cite{everingham2010pascal, lin2014microsoft}. However, FSOD requires instance-level annotated datasets that are expensive to obtain. Weakly-supervised object detection (WSOD) \cite{bilen2016weakly, tang2017multiple, diba2017weakly, zhang2018zigzag, wan2018min} is thus studied as it only needs image-level annotations. However, WSOD fails to achieve a satisfying result compared to fully-supervised object detection methods, which stimulates the demand for SSOD.

\textbf{Semi-supervised learning} trains models with both labeled and unlabeled data. Many works for semi-supervised learning appears in recent years, such as consistency regularization based methods \cite{laine2016temporal, miyato2018virtual, tarvainen2017mean, ouali2020semi, li2020transformation}, self-training \cite{yarowsky1995unsupervised, chapelle2009semi, peng2020deep}, label propagation \cite{zhu2002learning, bengio11label}, data augmentation \cite{xie2019unsupervised, berthelot2019mixmatch}, or entropy regularization \cite{kalluri2019universal}. Although semi-supervised learning develops rapidly, it usually targets at classification or semantic segmentation problems rather than detection. Object detection is in nature much more difficult so off-the-shell semi-supervised methods can hardly be directly applied to object detection.

\textbf{Semi-supervised object detection (SSOD)} aims to train a detector with both instance-level annotated data and unlabeled data. For example, \cite{hoffman2014lsda} utilizes unlabeled data to expand the number of categories that the detector recognizes, and \cite{zoph2020rethinking} studies the effect of self-training and pre-training. Different from them, we mainly utilize unlabeled images to boost the performance of FSOD. Current SSOD methods are generally divided into two groups. The first is based on pseudo labels \cite{wang2018towards, wang2018cost, radosavovic2018data}. They usually perform SSOD training only once and are susceptiable to the label noise overfitting problem. The second is to borrow ideas from semi-supervised classification methods and use consistency regularization \cite{jeong2019consistency, jeong2020interpolation}. But they are more suitable for one-stage detectors and behave poorly in two-stage models. Our approach is established on pseudo labels.

\section{Method}

\begin{figure}[]
   \setlength{\abovecaptionskip}{5pt}
   \setlength{\belowcaptionskip}{5pt}
\subfloat[]{ \includegraphics[width=0.45\columnwidth]{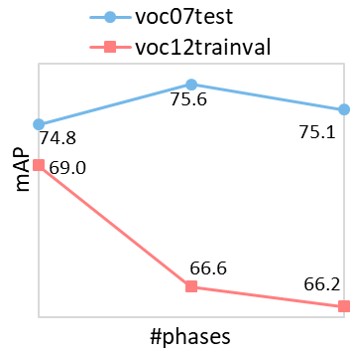} \label{fig:21}  }
     \hspace{0.04\columnwidth}
\raggedright \subfloat[]{ \includegraphics[width=0.45\columnwidth]{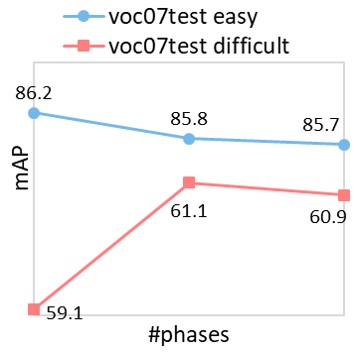}  \label{fig:easyd} }
\caption{ \textbf{Illustration of the label noise overfitting problem for detection models.} The data sequence is: supervised model - semi-supervised model from the first phase - semi-supervised model from the second phase. }
\label{fig:ofd}
\end{figure}

\begin{figure*}[]
\centering
   \setlength{\abovecaptionskip}{5pt}
   \setlength{\belowcaptionskip}{0pt}
      \includegraphics[width=2\columnwidth]{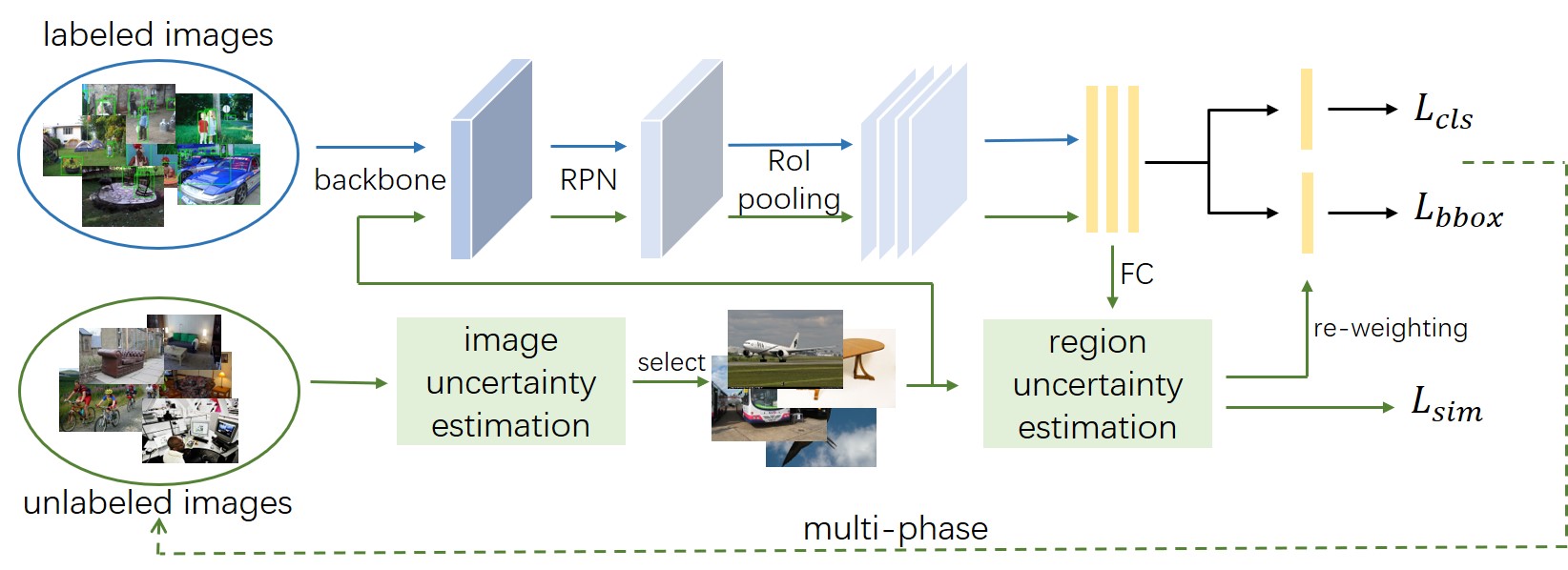}

\caption{\textbf{The diagram of uncertainty guided multi-phase learning.} Multi-phase self-training is designed for unlabeled images flow in SSOD. Image uncertainty estimation and region uncertainty estimation guide the multi-phase SSOD learning. }
\label{fig:bigframe}
\end{figure*}

\subsection{Label Noise Overfitting Problem\label{nl}}

For pseudo label based SSOD, pseudo labels are regarded as groundtruth annotations of unlabeled images for training the detector. Label noise inherently exists in these pseudo labels and brings about the uncertainty for SSOD training. We observe that deep learning detection models are susceptible to overfit to noisy labels, which damages the SSOD training. For further verification, we implement self-training SSOD twice with VOC 2007 trainval as labeled set and VOC 2012 trainval as unlabeled set. The performance such as mAP versus training phases is plotted in Fig. \ref{fig:21}.

From the empirical analysis we can see that the mAP on VOC07 test improves after the first SSOD phase because of the utility of extra unlabeled data, but it declines on VOC12 trainval. As a consequence, the quality of generated pseudo labels descends, causing worse test mAP in the second phase. This demonstrates that during SSOD training, the generalization ability of the model on testing data is strengthened while its ability on training data is weakened. This phenomenon derives from noisy pseudo labels. During SSOD training, noisy pseudo labels deliver uncertain supervised signals to the detectors, which attracts excessive attention. SSOD models attempt to follow this incorrect information, thus fitting to noisy labels and amplifying the noise. We refer to this as the label noise overfitting problem.

At image level, for an unlabeled image with pseudo labels, it consists of some correct knowledge that can improve the performance and some incorrect noise that hurts the training. If the correct knowledge is more, this image will benefit the model. We call this kind of image an easy one and vice versa. We propose that the recall of pseudo labels reflects the correct information it contains and 1-precision is a metric of noise. With this metric, we evaluate previous models on easy or difficult images from VOC07 test separately. From Fig. \ref{fig:easyd}, we observe that although the SSOD model's prediction on the test set is more precise, it deteriorates on easy images. Higher mAP on test set mainly comes from difficult images, suggesting that the model over-focuses on difficult images with more noise and ignores easy images, a consequence of the label noise overfitting problem. Therefore, this model is deficient in utilizing unlabeled images adequately. 

At region level, even for an easy image defined above where pseudo annotations are relatively clean, some objects within it may still lack pseudo annotations. In this circumstance, some positive regions are labeled as background category during training. The detector's classification module reveals that these regions are similar to some existing objects but they are not highly overlapped with any positive instances. This contradiction leads to the uncertainty of these regions. These noisy regions usually hold a large loss value thus are dominant over other regions during training, which negatively affects the performance.

\subsection{Multi-Phase Learning}

\begin{algorithm}[t]
\caption{The overall procedure for multi-phase SSOD learning.}
\begin{algorithmic}
\State \hspace*{-0.18in} \textbf{Require}:
\State The number of training phases, $N$
\State \hspace*{-0.18in} \textbf{Training}:
\State Train a FSOD model with all labeled data.
\State Set the initial easy data fraction: $k=1/N$
\For{$i=1$; $i<=N$; $i++$}
\State \hspace*{-0.25in} 1. Predict on unlabeled data with all current models.
\State \hspace*{-0.25in} 2. Take the intersection for all current pseudo labels.
\State \hspace*{-0.25in} 3. Select top $k$ easy images from unlabeled images. 
\State \hspace*{-0.25in} 4. Train a SSOD model with labeled and easy unlabeled data.
\State \hspace*{-0.25in} 5. $k = k + 1/N$
\EndFor

\State \hspace*{-0.18in} \textbf{Testing}:
\State Ensemble testing results from all models to generate ultimate results.
\end{algorithmic}
\label{alg:process}
\end{algorithm}

Current SSOD methods usually utilize unlabeled images once. Because of the label noise overfitting problem, difficult images with more noise are attached with higher importance and easy images are relatively discounted. One-phase learning generating a single SSOD model is hard to alleviate this problem. Because no matter how advanced the initial supervised detector is, pseudo labels of difficult images always accommodate more noise and dominate during the training. We thus leverage more than one model for easy or difficult data separately. Specifically, we choose easy unlabeled images to conduct SSOD first. In this training process, most of the data are relatively easy so the model will fit high-confident easy data well. Difficult images are added to train other subsequent detection models, and those models progressively concentrate on difficult images. Finally, we obtain a series of models that excel in different difficulty levels of images, and all of them together can fit all unlabeled data. During inference, we consider all models to fully exploit different abilities of divergent models. We use weighted boxes fusion \cite{solovyev2019weighted} to ensemble detection results from all models. In this way, different types of information are utilized comprehensively. 

For beginning models, they are trained with easy images thus are less affected by the label noise overfitting problem. With stronger generalization ability and milder overfitting to noisy labels, their predictions on the unlabeled training set are more convincing. Their generated pseudo labels can be provided for later training without performance reduction appearing in previous self-training methods. The above training ultimately forms a multi-phase procedure.

For a particular training phase, it is more appropriate to synthesize all models from previous phases for creating new pseudo labels, rather than just depend on the latest model. We consider the \emph{intersection} of pseudo labels from all previous models. Intersection operation improves the precision and further reduces the uncertainty of pseudo labels. After the intersection, all models reach a consensus on each pseudo annotation, resulting in higher self-confidence and certainty. The overall process is in Algorithm \ref{alg:process}.

\subsection{Uncertainty Guided Training}

\subsubsection{Image Uncertainty Guided Selection}

\quad To proceed with our method, we need to select easy images from the unlabeled dataset. Since annotations of unlabeled images are inaccessible, the above recall/precision metric for discriminating whether an image is easy is unavailable. We need an alternative metric that should guarantee that the detection model is more certain about easy images than difficult ones. Now that we have detected objects (\emph{i.e.} pseudo labels) $\left\{ (bb_{mn}, s_{mn}) \right\}_{m=1}^M$ of Image $I_m$, where $bb_{mn}$ denotes the bounding box and $s_{mn}$ is the corresponding confidence score, $s_{mn}$ is a reasonable and simple representation of how certain the model is about the specific object. The average of all bounding boxes' scores inside an image measures the certainty degree of all annotations. This corresponds with the image's difficulty, just as follows:

\begin{equation}
    \overline{s}_m = \sum_{m=1}^Ms_{mn}/M
\label{equ:diff}
\end{equation}

The above formulation provides a description of image uncertainty. It selects images that detectors are more certain about and enables that detection models are certain about each annotation in pseudo labels. Images with a small $\overline{s}_m$ are regarded as difficult ones and are filtered out in the first several phases. This selection strategy based on image uncertainty guides the detector on more certain images to mitigate the label noise overfitting problem.

\vspace{-0.8em}

\subsubsection{Region Uncertainty Guided RoI Re-weighting}

\begin{figure}[t]
   \centering
   \setlength{\abovecaptionskip}{5pt}
   \setlength{\belowcaptionskip}{5pt}
   \includegraphics[width=\columnwidth]{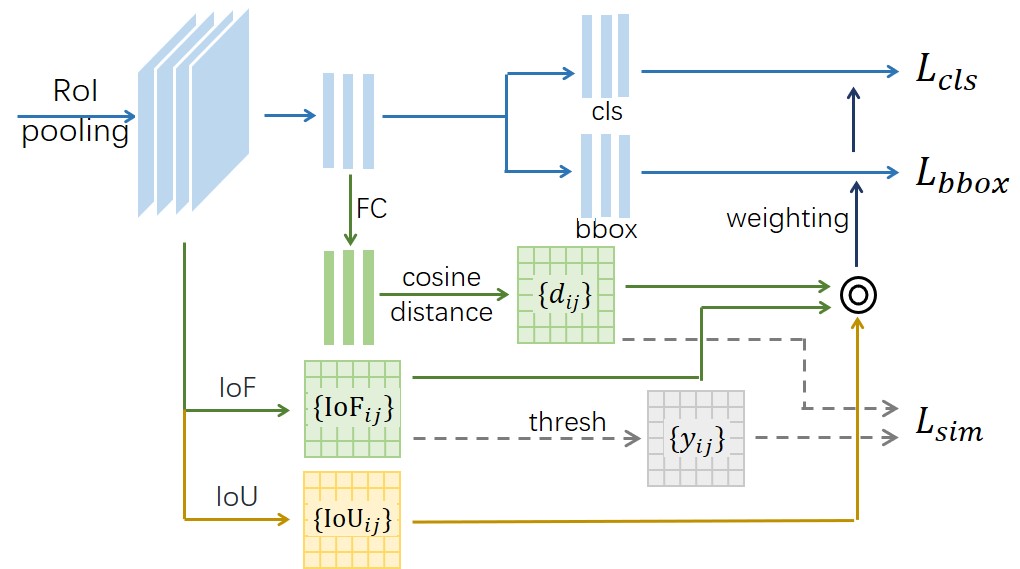}
   \caption{\textbf{The pipeline of uncertainty based re-weighting.} Similarity, IoUs and IoFs among different objects are calculated for weights, and IoFs are treated as groundtruth labels for similarity learning. }
   \label{fig:sss}
\end{figure}

\quad With the above training framework, we exclude difficult images in the initial phases and integrate pseudo labels from different models by intersection. SSOD is hence able to avoid uncertain images with more noise, especially in the initial phases. But the detector is still distracted by noisy regions with missing annotation problems. To alleviate this, we introduce our uncertainty based re-weighting strategy. Details are shown in Fig. \ref{fig:sss}. Concretely, we illustrate region-level uncertainty and re-weight for RoIs with their uncertainty guided during training. The strategy discovers uncertain regions and reduces their gradients by down-weighting to facilitate more accurate and certain regions standing out.

According to \cite{wu2018soft}, background RoIs which are hardly overlapped with all positive instances are more likely to be miss-annotated. Based on it, we adopt Intersection over Union (IoU) as one of the metrics for uncertainty measurement. Overlap based weights are computed as follows:

\begin{equation}
    w_i = a + (1-a)e^{-be^{-c \cdot {\rm IoU_i}}}
    \label{eqa:ss}
\end{equation}
where $a,b,c$ are pre-defined parameters, IoU$_i$ is the maximum IoU between the negative RoI$_i$ and all positive RoIs. $w_i$ is the reduced weight for RoI$_i$.

If a region is attributed with high similarity with some current objects but is not pseudo-labeled, we can claim that it is uncertain. Besides of IoU, the similarity between different RoIs should be a necessary description of uncertainty. We apply the cosine distance to calculate the similarity between different RoIs. Let $f_i$ denote the feature of RoI$_i$. The distance (\emph{i.e.} dissimilarity) $d_{ij}$ between RoI$_i$ and RoI$_j$ is:

\begin{equation}
    d_{ij} = \frac{ | f_i^Tf_j |} {\Vert f_i \Vert \Vert f_j \Vert}
\end{equation}

We notice that there are many small RoIs inside a large RoI after RPN. These small RoIs are usually highly similar to the large one, but they are not uncertain because their categories should be background, not other positive ones. Intersection over foreground (IoF) can quantify this phenomenon. We combine both cosine distance and IoF for ultimate similarity based uncertainty description:

\begin{equation}
    D_i = 1 - \max_j d_{ij}(1-{\rm IoF_{ij}})
\end{equation}

The ultimate uncertainty based weights are as follows:

\begin{equation}
    w_i = (a + (1-a)e^{-be^{-c_1 \cdot IoU_i}})e^{-be^{-c_2D_i}}
    \label{eqa:w}
\end{equation}
where $a,b,c_1, c_2$ are pre-defined parameters.

Equation \ref{eqa:w} is composed of two items, stemming from overlap based uncertainty and similarity based uncertainty separately. The product offers a balance between two different uncertainty. This formulation is based on the Gompertz function, a special form of the logistic function, where $b$ is a large value. If the region is uncertain (IoU$_i$ and $D_i$ are small), its weight is close to 0, and its uncertain gradient is diminished to prevent from back-propagating through the network. As the uncertainty degree decreases, the weight increases rapidly to 1 for normal training.

But noisy knowledge still impairs the above process since raw RoI features are supervised with noisy labels in SSOD models. To further avoid noisy missing annotation labels, we embed the feature for similarity with a fully-connected layer. The purpose is to decrease distances among the same objects and increase distances among different objects. Considering that a bounding box is highly similar to boxes inside it, we adopt IoF as a supervised signal for similarity learning:

\begin{equation}
    y_{ij} = I({\rm IoF_{ij}}>t) \ or \ I({\rm IoF_{ji}}>t)
\end{equation}
where $I(\cdot)$ is the indicator function, $t$ is a pre-defined thresh. We set $t$ to 0.7 in all experiments. Ultimately, the loss function for learning similarity is defined as

\begin{equation}
    L_{sim} = {y_{ij}(1-d_{ij})^2 + (1-y_{ij})d_{ij}^2 }
\end{equation}

\section{Experiments}

We evaluate the proposed method for SSOD on PASCAL VOC \cite{everingham2010pascal} and MSCOCO \cite{lin2014microsoft}. For VOC, we use VOC 2007 trainval (5,011 images) as the labeled data and VOC 2012 trainval (11,540 images) as the unlabeled data, then evaluate on VOC 2007 test (4,952 images). For COCO, we refer to a 35k subset of COCO 2014 validation set as co-35, the 80k training set as co-80 and the union of them as co-115. 120k unlabeled images from COCO 2017 is called as co-120. We use two settings for semi-supervised training: 1) co-35 as labeled set, co-80 as unlabeled set; 2) co-115 as labeled set, co-120 as unlabeled set, then report the model performance on COCO 2014 minival set (5,000 images). All experiments are implemented with PyTorch \cite{paszke2019pytorch} and MMDetection \cite{chen2019mmdetection}. Pseudo labels are obtained from the model's predictions post-processed by a per-category threshold as in \cite{radosavovic2018data}. Unless otherwise specified, we use ResNet50 \cite{he2016deep} based Faster RCNN \cite{ren2015faster} for two-phase training. All other experimental settings are the same as MMDetection.

\subsection{Comparison with Existing Methods}

\textbf{PASCAL VOC.} We perform the comparative study for SSOD based on the two-stage detector Faster RCNN and one-stage detector SSD \cite{liu2016ssd} on VOC dataset. The results are presented in Tab. \ref{tab:voc}. For Faster RCNN, we note that our method significantly outperforms the previous one-phase SSOD baseline model, which improves the mAP of the fully-supervised model (FS) on VOC07 from 74.8\% to 75.6\%. For comparison, our data uncertainty-based multi-phase learning achieves the mAP of 78.6\% and increases the mAP by 3\% compared to the baseline. Compared to DD \cite{radosavovic2018data} which produces more accurate pseudo labels, the mAP is increased by 2.6\% even with relatively low-quality pseudo labels. The experimental results show that our method is quite efficacious for SSOD. The less than 3\% gap between our method (78.6\%) and the upper bound obtained by fully-supervised learning on VOC0712 (81.2\%) demonstrates the strong ability of our method to learn the knowledge within unlabeled data. 

For SSD \cite{liu2016ssd}, our uncertainty-based multi-phase learning achieves the mAP of 74.5\%, 2.7\% higher than the baseline of one-phase learning, Compared to consistency-based semi-supervised learning (CSD) \cite{jeong2019consistency} and interpolation-based semi-supervised learning (ISD) \cite{jeong2020interpolation}, the detection mAP is improved by 2.2\% and 1.2\% respectively. However, these two methods behave poorly when combined with two-stage detectors. In contrast, our method works consistently well for both two-stage and one-stage detectors.

\begin{table}[]
\centering
\setlength{\abovecaptionskip}{0pt}
\setlength{\belowcaptionskip}{0pt}
\caption{\textbf{Semi-supervised Detection Results on PASCAL VOC 2007 test} \emph{vs.} current SSOD methods and fully-supervised results trained on VOC07 or VOC0712. (L: labeled data, Un: unlabeled data.)}
\resizebox{\columnwidth}{!}{
\begin{tabular}{c|c|c|cc|c}
\Xhline{1.2pt}
Model & Backbone & Method & \tabincell{c}{L} & \tabincell{c}{Un} & mAP \\
\hline

\multirow{5}{1.2cm}{Faster RCNN}  & \multirow{5}{1.2cm}{ResNet50} & FS & VOC07 & - & 74.8  \\
 & & Baseline & VOC07 & VOC12 & 75.6\\
 & & DD \cite{radosavovic2018data} & VOC07 & VOC12 & 76.0\\
 & & ours & VOC07 & VOC12 & \textbf{78.6}\\
 & & FS & VOC0712 & - & 81.2\\

\hline
\multirow{6}{1.2cm}{SSD300}  & \multirow{6}{1.2cm}{VGG16} & FS & VOC07 & - & 70.2\\ 
 & & Baseline & VOC07 & VOC12 & 71.8\\
 & & CSD \cite{jeong2019consistency} & VOC07 & VOC12 & 72.3\\
 & & ISD \cite{jeong2020interpolation} & VOC07 & VOC12 & 73.3\\
 & & ours & VOC07 & VOC12 & \textbf{74.5}\\ 
 & & FS & VOC0712 & - & 77.2\\

\Xhline{1.2pt} 
\end{tabular}
}
\label{tab:voc}
\end{table}

\begin{table}[]
\centering
\setlength{\abovecaptionskip}{0pt}
\setlength{\belowcaptionskip}{0pt}
\caption{\textbf{Semi-supervised detection Results on COCO minival} \emph{vs.} current SSOD and FSOD results. $^\dag$ denotes that the performance is obtained by the final model after the multi-phase learning without ensemble.}
\resizebox{\columnwidth}{!}{
\begin{tabular}{c|c|cc|ccc}
\Xhline{1.2pt}
Backbone & Method & \tabincell{c}{L} & \tabincell{c}{Un} & $AP$ & $AP_{50}$ & $AP_{75}$\\
\hline
\multirow{10}{1.3cm}{ResNet50} & FS & co-35 & - & 31.3 & 52.0 & 33.0\\
  & DD & co-35 & co-80 & 33.1 & 53.3 & 35.4 \\
  & ours & co-35 & co-80 & \textbf{34.8} & \textbf{55.1} & \textbf{37.2} \\
  & ours + DD & co-35 & co-80 & \textbf{35.2} & \textbf{55.7} & \textbf{37.6}\\
  \cline{2-7}
  & FS & co-115 & - & 37.4 & 58.1 & 40.4\\
  & DD & co-115 & co-120 & 37.9 & 60.1 & 40.8 \\
  & PL \cite{tang2020proposal} & co-115 & co-120 & 38.4 & 59.7 & 41.7 \\
  & ours & co-115 & co-120 & \textbf{40.1} & \textbf{60.4} & \textbf{43.7} \\
  & ours$^\dag$ + DD & co-115 & co-120 & 38.9 & 59.4 & 42.3 \\
  & ours + DD & co-115 & co-120 & \textbf{40.3} & \textbf{61.0} & \textbf{43.9}\\
\hline
\multirow{5}{1.3cm}{ResNet101}  & FS & co-115 & - & 39.4 & 60.1 & 43.1\\
  & DD & co-115 & co-120 & 40.1 & 62.1 & 43.5 \\
  & ours & co-115 & co-120 & \textbf{42.2} & \textbf{62.5} & \textbf{46.1}\\
  & ours$^\dag$ + DD & co-115 & co-120 & 41.2 & 61.5 & 44.9 \\
  & ours + DD & co-115 & co-120 & \textbf{42.3} & \textbf{62.7} & \textbf{46.3}\\
\Xhline{1.2pt} 
\end{tabular}
}
\label{tab:coco}
\end{table}

\textbf{MS COCO}. We conduct the comparative experiments with Faster RCNN as the base detector on COCO dataset. The metrics to measure detection accuracy such as $AP$, $AP_{50}$, $AP_{75}$ are presented in Tab. \ref{tab:coco}. Note that our reproduced DD on co-35/80 performs a little better than the original paper. For co-35/80 split, the proposed multi-phase learning based on ResNet50 backbone achieves the $AP$ of 34.8\% for SSOD, outperforming DD by 1.7\%. Moreover, we combine our method with DD, achieving the $AP$ of 35.2\%. Compared to DD, the achieved $AP$ is increased by 2.1\%, which is extremely prominent for COCO dataset, especially in semi-supervised settings. For co-115/120 split, our method also achieves consistently better than DD method and the state-of-the-art method on coco-115/120 - proposal learning (PL) \cite{tang2020proposal}. Even without using ensemble to excavate knowledge of different difficulty levels, our final model still outperforms DD by 1\% and PL by 0.5\%. With ResNet101, a much more powerful feature extractor, the overall $AP$ is further enhanced to 42.3\%. The results validate the efficiency of our proposed multi-phase learning guided by data uncertainty and its strong ability to transcend the upper bound of traditional fully-supervised methods.

\subsection{Ablation Study}

\begin{table}[]
\centering
\setlength{\abovecaptionskip}{0pt}
\setlength{\belowcaptionskip}{0pt}
\caption{\textbf{Ablation Study on PASCAL VOC 2007 test.} (RR: RoI Re-weighting)}
\resizebox{\columnwidth}{!}{
\begin{tabular}{c|cc|ccc|c}
\Xhline{1.2pt}
Model & \tabincell{c}{L} & \tabincell{c}{Un} &  \tabincell{c}{Two-\\ Phase} & \tabincell{c}{RR}  & \tabincell{c}{Ensemble} & mAP \\
\hline

\multirow{5}{1.2cm}{Faster RCNN}  & VOC07 & - & & &  &    74.8  \\
  &    VOC07 & VOC12 & & &  & 75.6 \\
  &    VOC07 & VOC12 & \checkmark & &  & 76.1 \\
  &    VOC07 & VOC12 & \checkmark & \checkmark &  & 77.4 \\
  &    VOC07 & VOC12 & \checkmark & \checkmark & \checkmark & \textbf{78.6} \\
  
\hline 
\multirow{4}{1.2cm}{SSD300} & VOC07 & - & & &  &    70.2 \\
  &    VOC07 & VOC12 & & &  & 71.8 \\
  &    VOC07 & VOC12 & \checkmark & &  &  72.3\\
  &    VOC07 & VOC12 & \checkmark & & \checkmark  & \textbf{74.5} \\
\Xhline{1.2pt} 
\end{tabular}
}
\label{tab:abl}
\end{table}

We perform ablation study on PASCAL VOC to analyze the impacts of 1) multi-phase learning (two-phase concretely), 2) RoI Re-weighting strategy that enables the model to focus on certain regions, 3) model ensembling during inference. The results are in Tab. \ref{tab:abl}.

For Faster RCNN, two-phase training is 0.5\% higher than one-phase training. This improvement reveals that our method alleviates the label noise overfitting problem that causes performance deterioration. RoI re-weighting further produces a 1.5\% mAP gain, which confirms the ability of re-weighting to force the detector into more certain regions and reduce the missing annotation noise effect. Results ensembling finally boosts the mAP with 1.2\% improvement. According to previous studies \cite{xu2020multi}, model ensembling performs poorly when ensembled models are similar to each other. The significant promotion brought by ensembling in our method thereby indicates that models from different phases master in images with different difficulty levels.  

For SSD, since RoI re-weighting cannot be applied to one-stage detectors, the mAP gain is a little lower. It is also noticeable that ensembling contributes more to the SSD detector, resulting in 2.2\% mAP improvement. This is because one-stage detectors are usually uncompetitive in performance compared to two-stage detectors and could hardly excavate sufficient information. Models possess more random elements thus benefit more from ensembling.

\begin{table}[]
\centering
\setlength{\abovecaptionskip}{0pt}
\setlength{\belowcaptionskip}{0pt}
\caption{\textbf{Effect of RoI Re-weighting on SSOD compared with Baseline and Soft Sampling.}  ${0 \sim 2}$ is the ensemble result of model from phase 0 (FS model) to phase 2.}
\resizebox{\columnwidth}{!}{
\begin{tabular}{c|c|c|c}
\Xhline{1.2pt}
Phase &  Baseline & Soft Sampling & RoI Re-weighting \\
\hline
0 (FS Model) & 74.8 & 74.8 & 74.8\\
1 & 75.9 & 76.2 & \textbf{76.6} \\
2 & 76.1 & 76.6 & \textbf{77.4} \\
${0 \sim 2}$ & 77.8 & 78.1 & \textbf{78.6}\\

\Xhline{1.2pt} 
\end{tabular}
}
\label{tab:uwl}
\end{table}

\begin{figure}[]
   \centering
   \setlength{\abovecaptionskip}{5pt}
   \setlength{\belowcaptionskip}{5pt}
   \includegraphics[width=\columnwidth]{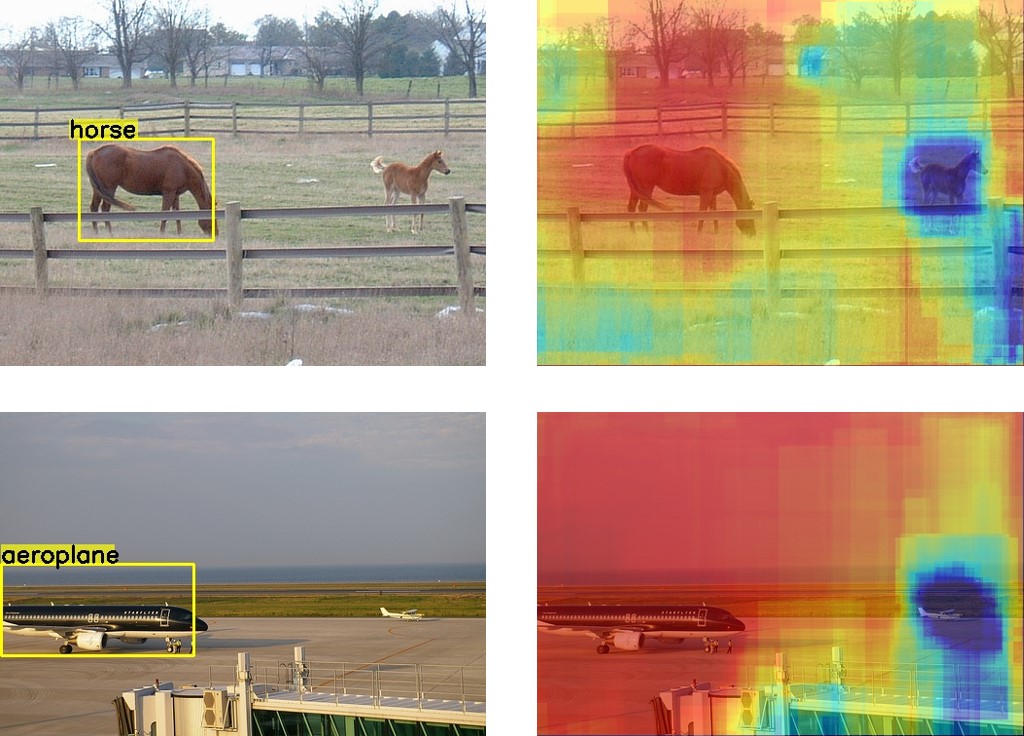}
   \caption{\textbf{Illustrative examples for RoI Re-weighting.} The left column is pseudo labels with missing annotations, and the right column is the heatmap for region uncertainty after RR. Blue regions are more uncertain for the detector. Before RR, all regions have the same weight of 1.0, while after RR, uncertain regions are assigned with lower weights. }
   \label{fig:uwlimages}
\end{figure}

\hspace*{-0.18in} $\bullet$ \textbf{RoI Re-weighting Analysis.} We further study the effect of RoI re-weighting. We compare our method with baseline and overlap based soft sampling \cite{wu2018soft}. The results are listed in Tab. \ref{tab:uwl}. For a raw Faster RCNN, mAP increases from 74.8\% to 75.9\% in the first phase but almost remains unchanged in the second phase. This is because Faster RCNN is easily misguided by noisy labels. For the first phase, unlabeled images are easy and pseudo labels are relatively clean. In such a situation, Faster RCNN can obtain a satisfying result. But in the second phase, difficult noisy unlabeled images participate in training so the performance is seriously affected. Soft sampling mitigates the missing annotation problem to some extent. But it regards region-level uncertainty metric just as a function of overlaps, which is insufficient. Take images in Fig. \ref{fig:uwlimages} for example. For the first one, the horse in the right is not labeled and for the second one, the tiny plane in the middle is missed. The common is that almost all background RoIs share little overlaps with positive instances and are down-weighted to the same value with just overlap based metric. For our method based on both similarity and overlaps metric, as shown in the right column in Fig. \ref{fig:uwlimages}, uncertain regions with missing annotation problems are successfully detected (blue regions) and their weights are further reduced. As a result, uncertain gradient information is depressed and our model is able to focus on more certain regions. Then the final model is more robust, especially in the second phase.

\begin{table}[]
\centering
\setlength{\abovecaptionskip}{0pt}
\setlength{\belowcaptionskip}{0pt}
\caption{ \textbf{Detection results from different models.} ${0 \sim 2}$ indicates the ensemble result of model from phase 0 (fully-supervised model) to phase 2.}
\resizebox{0.9\columnwidth}{!}{
\begin{tabular}{c|c|c|c}
\Xhline{1.2pt}
Phase &  \tabincell{c}{VOC07\\Test} & \tabincell{c}{VOC07 Test \\ (easy)} & \tabincell{c}{VOC07 Test \\ (difficult)}\\
\hline
0 (FS Model) & 74.8  & 86.2 & 59.1\\
1  &  76.6  & \textbf{86.7} & 62.7\\
2  &  \textbf{77.4}  & 86.4 & \textbf{63.8} \\
\hline
${1 \sim 2}$ &  78.3 & \textbf{87.3} & 65.4 \\
${0 \sim 2}$ &  \textbf{78.6}  & \textbf{87.3} & \textbf{66.2} \\
\Xhline{1.2pt} 
\end{tabular}
}
\label{tab:ensem}
\end{table}

\hspace*{-0.18in} $\bullet$ \textbf{Model Divergence Analysis.} We evaluate the models from multiple phases on easy or difficult images from the test set like section \ref{nl}. From Tab. \ref{tab:ensem}, we observe that the model from the first phase performs best for easy images, which indicates that the model trained with just easy unlabeled data fixates on certain knowledge. The model from the second phase has the best generalization ability since it learns with the largest amount of data. But it performs worse on easy data due to the label noise overfitting problem that makes the model attach great importance to difficult data and overlook easy data. This experiment verifies that models from different phases are experienced on data of different difficulty levels. Since these two models target at easy and difficult images separately, they are complementary to each other and ensembling them brings about a large improvement. We also notice that these two models are already able to fit all certain information thus adding the FS model does not boost performance on easy images. For difficult uncertain features, the FS model is still able to complement a little and enhance the final mAP.

\subsection{Discussion}

In previous experiments, we conduct two-phase SSOD training. In this section, we evaluate our method with more overall experiments to discuss the effect of learning process.

\subsubsection{Two-Phase Learning \label{tsp} }

\begin{figure}[]
   \centering
   \includegraphics[width=\columnwidth]{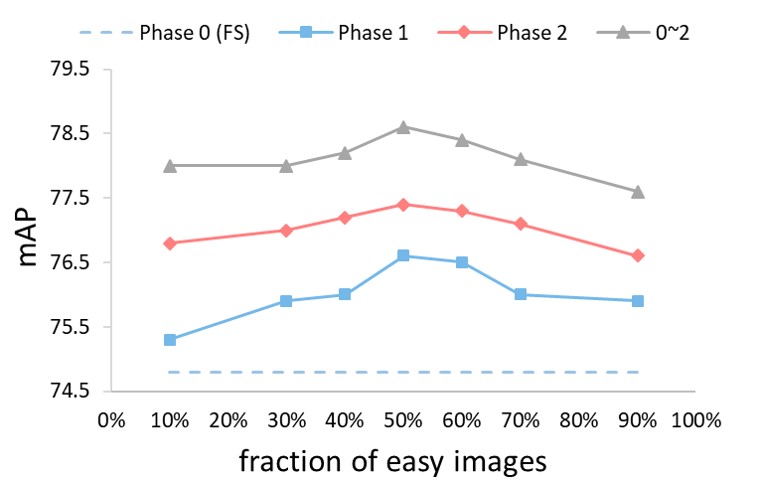}
   \caption{\textbf{Two-phase SSOD with different amount of easy data,} mAP reaches the peak when the ratio is 50\%. }
   \label{fig:twop}
\end{figure}

\quad According to Algorithm \ref{alg:process}, our semi-supervised learning increases the number of unlabeled images evenly in different phases. In this section, we perform two-phase SSOD with a variant amount of easy images for the first phase, then continue with all unlabeled images for the second phase. The results are plotted in Fig. \ref{fig:twop}. A positive correlation between the rate of easy images and the first phase mAP is observed when easy images are not too many, because more available unlabeled data contribute to stronger generalization ability. The performance attains the highest mAP when easy images are about 50\%, almost 76.8\% mAP. Then, the performance declines steadily when the percentage of easy images continues to increase, because noise within pseudo labels increases and label noise overfitting problem becomes severer as more difficult unlabeled data are accessible. The ability of the first phase model directly influences the quality of pseudo labels for the second phase. Ultimately, the mAP is the highest when easy images are about a half of total unlabeled images. Above results suggest that in SSOD settings, the volume of unlabeled data for training needs to be considered carefully. \textbf{More data do not necessarily lead to better performance for SSOD.}

Why results are the best when easy images are about 50\%? We evaluate pseudo labels from VOC and COCO with recall/precision metric presented in section \ref{nl} and find that the fraction of easy images is close to 50\% in both VOC and COCO. Even after data distillation \cite{radosavovic2018data}, easy data still occupy approximately 50\%. Easy images are those which contain more correct information than noise and contribute positively to SSOD training, so we can expect the best performance if all easy data participate in the first phase. For SSOD settings, the labeled dataset and unlabeled dataset share the same distribution and their quantity proportion is moderate. The labeled dataset is a little smaller, but is enough to train a nice supervised detection model. Images are also naturally collected. As a result, for an unlabeled image that the model has not ever seen before, the possibility that it is easy should be close to $50\%$. We thus assert that without any prior knowledge about the dataset, 50\% is a good estimation of easy image proportion and an appropriate value for implementation.

\subsubsection{More Phases \label{morestep}}

\begin{figure}[]
   \centering
   \includegraphics[width=\columnwidth]{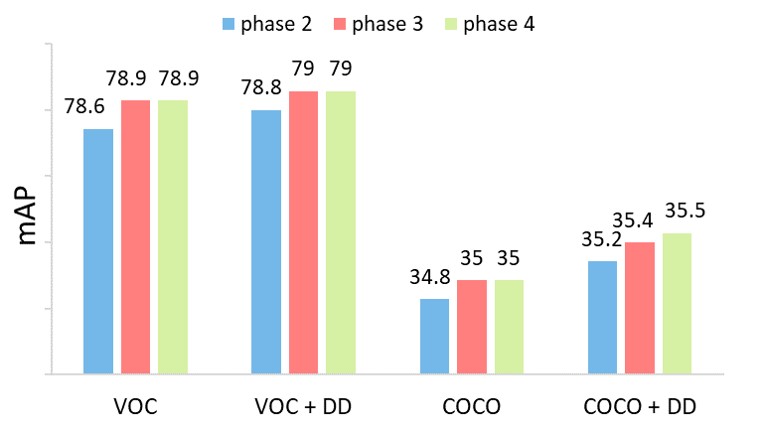}
   \caption{\textbf{Multiple phases semi-supervised learning on VOC07 test and COCO minival.} }
   \label{fig:morep}
\end{figure}

\quad We extend our method to more phases. Unlabeled data are evenly divided for different phases since datasets are naturally distributed. The results are in Fig. \ref{fig:morep}. For VOC dataset, the performance improves slightly from two-phase 78.6\% to three-phase 78.9\%, while four-phase learning does not produce better mAP. We believe two-phase learning that generates two semi-supervised models is already able to describe most unlabeled information adequately - one for easy certain information and the other for difficult uncertain information. Since information has been fully encoded, we do not need more models, \emph{i.e.}, more phases. For three-phase learning, the one more model may make up a little missing information. But when the phases continue to raise, existing models are already sufficient for all information and extra models cannot offer more information. For COCO dataset, although images are more complicated, two-phase is also sufficient and more phases do not introduce significant improvement. Therefore, we assert that two-phase learning is a good choice in practice.

\section{Conclusion}

In this paper, we propose a novel data uncertainty guided multi-phase learning method for semi-supervised object detection. The multi-phase training method enables the model to fully utilize all information and uncertainty descriptions guide the training process to make the detector concentrate on certain knowledge. We demonstrate the extraordinary ability of our method to excavate unlabeled knowledge and achieve state-of-the-art performance. Semi-supervised object detection is a challenging problem and we will further explore how to utilize unlabeled data more efficiently.

\section*{Acknowledgment}

\hspace*{-0.18in} This work was supported by the National Natural Science Foundation of China under Grant No.61771288, Cross-Media Intelligent Technology Project of Beijing National Research Center for Information Science and Technology (BNRist) under Grant No.BNR2019TD01022 and the research fund under Grant No. 2019GQG0001 from the Institute for Guo Qiang, Tsinghua University. We also thank for the funding support of Huawei Technologies Co. Ltd.

{\small
\bibliographystyle{ieee_fullname}
\bibliography{egbib}

\begin{thebibliography}{10}\itemsep=-1pt

\bibitem{arpit2017closer}
Devansh Arpit, Stanislaw~K Jastrzebski, Nicolas Ballas, David Krueger, Emmanuel
  Bengio, Maxinder~S Kanwal, Tegan Maharaj, Asja Fischer, Aaron~C Courville,
  Yoshua Bengio, et~al.
\newblock A closer look at memorization in deep networks.
\newblock In {\em ICML}, 2017.

\bibitem{bengio11label}
Yoshua Bengio, Olivier Delalleau, and Nicolas Le~Roux.
\newblock label propagation and quadratic criterion.(2006).
\newblock 11.

\bibitem{berthelot2019mixmatch}
David Berthelot, Nicholas Carlini, Ian Goodfellow, Nicolas Papernot, Avital
  Oliver, and Colin~A Raffel.
\newblock Mixmatch: A holistic approach to semi-supervised learning.
\newblock In {\em Advances in Neural Information Processing Systems}, pages
  5049--5059, 2019.

\bibitem{bilen2016weakly}
Hakan Bilen and Andrea Vedaldi.
\newblock Weakly supervised deep detection networks.
\newblock In {\em Proceedings of the IEEE Conference on Computer Vision and
  Pattern Recognition}, pages 2846--2854, 2016.

\bibitem{Cai_2019}
Zhaowei Cai and Nuno Vasconcelos.
\newblock Cascade r-cnn: High quality object detection and instance
  segmentation.
\newblock {\em IEEE Transactions on Pattern Analysis and Machine Intelligence},
  page 1–1, 2019.

\bibitem{chapelle2009semi}
Olivier Chapelle, Bernhard Scholkopf, and Alexander Zien.
\newblock Semi-supervised learning (chapelle, o. et al., eds.; 2006)[book
  reviews].
\newblock {\em IEEE Transactions on Neural Networks}, 20(3):542--542, 2009.

\bibitem{chen2019mmdetection}
Kai Chen, Jiaqi Wang, Jiangmiao Pang, Yuhang Cao, Yu Xiong, Xiaoxiao Li,
  Shuyang Sun, Wansen Feng, Ziwei Liu, Jiarui Xu, et~al.
\newblock Mmdetection: Open mmlab detection toolbox and benchmark.
\newblock {\em arXiv preprint arXiv:1906.07155}, 2019.

\bibitem{dai2016r}
Jifeng Dai, Yi Li, Kaiming He, and Jian Sun.
\newblock R-fcn: Object detection via region-based fully convolutional
  networks.
\newblock In {\em Advances in neural information processing systems}, pages
  379--387, 2016.

\bibitem{diba2017weakly}
Ali Diba, Vivek Sharma, Ali Pazandeh, Hamed Pirsiavash, and Luc Van~Gool.
\newblock Weakly supervised cascaded convolutional networks.
\newblock In {\em Proceedings of the IEEE conference on computer vision and
  pattern recognition}, pages 914--922, 2017.

\bibitem{everingham2010pascal}
Mark Everingham, Luc Van~Gool, Christopher~KI Williams, John Winn, and Andrew
  Zisserman.
\newblock The pascal visual object classes (voc) challenge.
\newblock {\em International journal of computer vision}, 88(2):303--338, 2010.

\bibitem{girshick2015fast}
Ross Girshick.
\newblock Fast r-cnn.
\newblock In {\em Proceedings of the IEEE international conference on computer
  vision}, pages 1440--1448, 2015.

\bibitem{girshick2014rich}
Ross Girshick, Jeff Donahue, Trevor Darrell, and Jitendra Malik.
\newblock Rich feature hierarchies for accurate object detection and semantic
  segmentation.
\newblock In {\em Proceedings of the IEEE conference on computer vision and
  pattern recognition}, pages 580--587, 2014.

\bibitem{he2017mask}
Kaiming He, Georgia Gkioxari, Piotr Doll{\'a}r, and Ross Girshick.
\newblock Mask r-cnn.
\newblock In {\em Proceedings of the IEEE international conference on computer
  vision}, pages 2961--2969, 2017.

\bibitem{he2016deep}
Kaiming He, Xiangyu Zhang, Shaoqing Ren, and Jian Sun.
\newblock Deep residual learning for image recognition.
\newblock In {\em Proceedings of the IEEE conference on computer vision and
  pattern recognition}, pages 770--778, 2016.

\bibitem{hoffman2014lsda}
Judy Hoffman, Sergio Guadarrama, Eric~S Tzeng, Ronghang Hu, Jeff Donahue, Ross
  Girshick, Trevor Darrell, and Kate Saenko.
\newblock Lsda: Large scale detection through adaptation.
\newblock In {\em Advances in Neural Information Processing Systems}, pages
  3536--3544, 2014.

\bibitem{jeong2019consistency}
Jisoo Jeong, Seungeui Lee, Jeesoo Kim, and Nojun Kwak.
\newblock Consistency-based semi-supervised learning for object detection.
\newblock In {\em Advances in neural information processing systems}, pages
  10759--10768, 2019.

\bibitem{jeong2020interpolation}
Jisoo Jeong, Vikas Verma, Minsung Hyun, Juho Kannala, and Nojun Kwak.
\newblock Interpolation-based semi-supervised learning for object detection.
\newblock {\em arXiv preprint arXiv:2006.02158}, 2020.

\bibitem{kalluri2019universal}
Tarun Kalluri, Girish Varma, Manmohan Chandraker, and CV Jawahar.
\newblock Universal semi-supervised semantic segmentation.
\newblock In {\em Proceedings of the IEEE International Conference on Computer
  Vision}, pages 5259--5270, 2019.

\bibitem{krizhevsky2012imagenet}
Alex Krizhevsky, Ilya Sutskever, and Geoffrey~E Hinton.
\newblock Imagenet classification with deep convolutional neural networks.
\newblock In {\em Advances in neural information processing systems}, pages
  1097--1105, 2012.

\bibitem{kuznetsova2018open}
Alina Kuznetsova, Hassan Rom, Neil Alldrin, Jasper Uijlings, Ivan Krasin, Jordi
  Pont-Tuset, Shahab Kamali, Stefan Popov, Matteo Malloci, Tom Duerig, et~al.
\newblock The open images dataset v4: Unified image classification, object
  detection, and visual relationship detection at scale.
\newblock {\em arXiv preprint arXiv:1811.00982}, 2018.

\bibitem{laine2016temporal}
Samuli Laine and Timo Aila.
\newblock Temporal ensembling for semi-supervised learning.
\newblock {\em arXiv preprint arXiv:1610.02242}, 2016.

\bibitem{lecun1998gradient}
Yann LeCun, L{\'e}on Bottou, Yoshua Bengio, and Patrick Haffner.
\newblock Gradient-based learning applied to document recognition.
\newblock {\em Proceedings of the IEEE}, 86(11):2278--2324, 1998.

\bibitem{lee2013pseudo}
Dong-Hyun Lee.
\newblock Pseudo-label: The simple and efficient semi-supervised learning
  method for deep neural networks.
\newblock In {\em Workshop on challenges in representation learning, ICML},
  volume~3, 2013.

\bibitem{li2020transformation}
Xiaomeng Li, Lequan Yu, Hao Chen, Chi-Wing Fu, Lei Xing, and Pheng-Ann Heng.
\newblock Transformation-consistent self-ensembling model for semisupervised
  medical image segmentation.
\newblock {\em IEEE Transactions on Neural Networks and Learning Systems},
  2020.

\bibitem{lin2017focal}
Tsung-Yi Lin, Priya Goyal, Ross Girshick, Kaiming He, and Piotr Doll{\'a}r.
\newblock Focal loss for dense object detection.
\newblock In {\em Proceedings of the IEEE international conference on computer
  vision}, 2017.

\bibitem{lin2014microsoft}
Tsung-Yi Lin, Michael Maire, Serge Belongie, James Hays, Pietro Perona, Deva
  Ramanan, Piotr Doll{\'a}r, and C~Lawrence Zitnick.
\newblock Microsoft coco: Common objects in context.
\newblock In {\em European conference on computer vision}, pages 740--755.
  Springer, 2014.

\bibitem{liu2016ssd}
Wei Liu, Dragomir Anguelov, Dumitru Erhan, Christian Szegedy, Scott Reed,
  Cheng-Yang Fu, and Alexander~C Berg.
\newblock Ssd: Single shot multibox detector.
\newblock In {\em European conference on computer vision}, pages 21--37.
  Springer, 2016.

\bibitem{ma2018dimensionality}
Xingjun Ma, Yisen Wang, Michael~E Houle, Shuo Zhou, Sarah Erfani, Shutao Xia,
  Sudanthi Wijewickrema, and James Bailey.
\newblock Dimensionality-driven learning with noisy labels.
\newblock In {\em International Conference on Machine Learning}, pages
  3355--3364, 2018.

\bibitem{miyato2018virtual}
Takeru Miyato, Shin-ichi Maeda, Masanori Koyama, and Shin Ishii.
\newblock Virtual adversarial training: a regularization method for supervised
  and semi-supervised learning.
\newblock {\em IEEE transactions on pattern analysis and machine intelligence},
  41(8):1979--1993, 2018.

\bibitem{ouali2020semi}
Yassine Ouali, C{\'e}line Hudelot, and Myriam Tami.
\newblock Semi-supervised semantic segmentation with cross-consistency
  training.
\newblock In {\em Proceedings of the IEEE/CVF Conference on Computer Vision and
  Pattern Recognition}, pages 12674--12684, 2020.

\bibitem{paszke2019pytorch}
Adam Paszke, Sam Gross, Francisco Massa, Adam Lerer, James Bradbury, Gregory
  Chanan, Trevor Killeen, Zeming Lin, Natalia Gimelshein, Luca Antiga, et~al.
\newblock Pytorch: An imperative style, high-performance deep learning library.
\newblock In {\em Advances in neural information processing systems}, pages
  8026--8037, 2019.

\bibitem{peng2020deep}
Jizong Peng, Guillermo Estrada, Marco Pedersoli, and Christian Desrosiers.
\newblock Deep co-training for semi-supervised image segmentation.
\newblock {\em Pattern Recognition}, page 107269, 2020.

\bibitem{radosavovic2018data}
Ilija Radosavovic, Piotr Doll{\'a}r, Ross Girshick, Georgia Gkioxari, and
  Kaiming He.
\newblock Data distillation: Towards omni-supervised learning.
\newblock In {\em Proceedings of the IEEE conference on computer vision and
  pattern recognition}, pages 4119--4128, 2018.

\bibitem{redmon2016you}
Joseph Redmon, Santosh Divvala, Ross Girshick, and Ali Farhadi.
\newblock You only look once: Unified, real-time object detection.
\newblock In {\em Proceedings of the IEEE conference on computer vision and
  pattern recognition}, pages 779--788, 2016.

\bibitem{redmon2018yolov3}
Joseph Redmon and Ali Farhadi.
\newblock Yolov3: An incremental improvement, 2018.

\bibitem{ren2015faster}
Shaoqing Ren, Kaiming He, Ross Girshick, and Jian Sun.
\newblock Faster r-cnn: Towards real-time object detection with region proposal
  networks.
\newblock In {\em Advances in neural information processing systems}, pages
  91--99, 2015.

\bibitem{rosenberg2005semi}
Chuck Rosenberg, Martial Hebert, and Henry Schneiderman.
\newblock Semi-supervised self-training of object detection models.
\newblock 2005.

\bibitem{shrivastava2016training}
Abhinav Shrivastava, Abhinav Gupta, and Ross Girshick.
\newblock Training region-based object detectors with online hard example
  mining.
\newblock In {\em Proceedings of the IEEE conference on computer vision and
  pattern recognition}, pages 761--769, 2016.

\bibitem{solovyev2019weighted}
Roman Solovyev and Weimin Wang.
\newblock Weighted boxes fusion: ensembling boxes for object detection models.
\newblock {\em arXiv preprint arXiv:1910.13302}, 2019.

\bibitem{tang2020proposal}
Peng Tang, Chetan Ramaiah, Ran Xu, and Caiming Xiong.
\newblock Proposal learning for semi-supervised object detection.
\newblock {\em arXiv preprint arXiv:2001.05086}, 2020.

\bibitem{tang2017multiple}
Peng Tang, Xinggang Wang, Xiang Bai, and Wenyu Liu.
\newblock Multiple instance detection network with online instance classifier
  refinement.
\newblock In {\em Proceedings of the IEEE Conference on Computer Vision and
  Pattern Recognition}, pages 2843--2851, 2017.

\bibitem{tarvainen2017mean}
Antti Tarvainen and Harri Valpola.
\newblock Mean teachers are better role models: Weight-averaged consistency
  targets improve semi-supervised deep learning results.
\newblock In {\em Advances in neural information processing systems}, pages
  1195--1204, 2017.

\bibitem{wan2018min}
Fang Wan, Pengxu Wei, Jianbin Jiao, Zhenjun Han, and Qixiang Ye.
\newblock Min-entropy latent model for weakly supervised object detection.
\newblock In {\em Proceedings of the IEEE Conference on Computer Vision and
  Pattern Recognition}, pages 1297--1306, 2018.

\bibitem{wang2018cost}
Keze Wang, Liang Lin, Xiaopeng Yan, Ziliang Chen, Dongyu Zhang, and Lei Zhang.
\newblock Cost-effective object detection: Active sample mining with switchable
  selection criteria.
\newblock {\em IEEE transactions on neural networks and learning systems},
  30(3):834--850, 2018.

\bibitem{wang2018towards}
Keze Wang, Xiaopeng Yan, Dongyu Zhang, Lei Zhang, and Liang Lin.
\newblock Towards human-machine cooperation: Self-supervised sample mining for
  object detection.
\newblock In {\em Proceedings of the IEEE Conference on Computer Vision and
  Pattern Recognition}, pages 1605--1613, 2018.

\bibitem{wu2018soft}
Zhe Wu, Navaneeth Bodla, Bharat Singh, Mahyar Najibi, Rama Chellappa, and
  Larry~S Davis.
\newblock Soft sampling for robust object detection.
\newblock {\em arXiv preprint arXiv:1806.06986}, 2018.

\bibitem{xie2019unsupervised}
Qizhe Xie, Zihang Dai, Eduard Hovy, Minh-Thang Luong, and Quoc~V Le.
\newblock Unsupervised data augmentation for consistency training.
\newblock {\em arXiv preprint arXiv:1904.12848}, 2019.

\bibitem{xu2020multi}
Jie Xu, Wei Wang, Hanyuan Wang, and Jinhong Guo.
\newblock Multi-model ensemble with rich spatial information for object
  detection.
\newblock {\em Pattern Recognition}, 99:107098, 2020.

\bibitem{yarowsky1995unsupervised}
David Yarowsky.
\newblock Unsupervised word sense disambiguation rivaling supervised methods.
\newblock In {\em 33rd annual meeting of the association for computational
  linguistics}, pages 189--196, 1995.

\bibitem{zhang2016understanding}
Chiyuan Zhang, Samy Bengio, Moritz Hardt, Benjamin Recht, and Oriol Vinyals.
\newblock Understanding deep learning requires rethinking generalization.
\newblock {\em arXiv preprint arXiv:1611.03530}, 2016.

\bibitem{DynamicRCNN}
Hongkai Zhang, Hong Chang, Bingpeng Ma, Naiyan Wang, and Xilin Chen.
\newblock Dynamic {R-CNN}: Towards high quality object detection via dynamic
  training.
\newblock {\em arXiv preprint arXiv:2004.06002}, 2020.

\bibitem{zhang2018zigzag}
Xiaopeng Zhang, Jiashi Feng, Hongkai Xiong, and Qi Tian.
\newblock Zigzag learning for weakly supervised object detection.
\newblock In {\em Proceedings of the IEEE Conference on Computer Vision and
  Pattern Recognition}, pages 4262--4270, 2018.

\bibitem{zhu2002learning}
Xiaojin Zhu and Zoubin Ghahramani.
\newblock Learning from labeled and unlabeled data with label propagation.
\newblock 2002.

\bibitem{zoph2020rethinking}
Barret Zoph, Golnaz Ghiasi, Tsung-Yi Lin, Yin Cui, Hanxiao Liu, Ekin~D Cubuk,
  and Quoc~V Le.
\newblock Rethinking pre-training and self-training.
\newblock {\em arXiv preprint arXiv:2006.06882}, 2020.

\end{thebibliography}
}

\end{document}